# Rules still work for Open Information Extraction

Jialin Hua[a, *], Liangqing Luo[a], Weiying Ping[a], Yan Liao[a, *], Chunhai Tao[a], Xuewen Lu[b]

[a]Key Laboratory of Data Science in Finance and Economics, and School of Statistics and Data Science, Jiangxi University of Finance and Economics, Nanchang, China

[*] Corresponding authors: nhma0004@gmail.com (Jialin Hua)

[b]Department of Mathematics and Statistics, University of Calgary, Calgary, Canada

[*] Corresponding authors: ly8189@126.com (Yan Liao)

**Abstract:** Open information extraction (OIE) aims to extract surface relations and their corresponding arguments from natural language text, irrespective of domain. This paper presents an innovative OIE model, APRCOIE, tailored for Chinese text. Diverging from previous models, our model generates extraction patterns autonomously. The model defines a new pattern form for Chinese OIE and proposes an automated pattern generation methodology. In that way, the model can handle a wide array of complex and diverse Chinese grammatical phenomena. We design a preliminary filter based on tensor computing to conduct the extraction procedure efficiently. To train the model, we manually annotated a large-scale Chinese OIE dataset. In the comparative evaluation, we demonstrate that APRCOIE outperforms state-of-the-art Chinese OIE models and significantly expands the boundaries of achievable OIE performance. The code of APRCOIE and the annotated dataset are released on GitHub (https://github.com/jialin666/APRCOIE_v1)

## 1. Introduction

The rapid growth of digital information has led to an overwhelming amount of text data in human languages, necessitating efficient methods to extract meaningful information. Being fundamentally designed to process structured data, computers encounter significant challenges when confronted with unstructured human language-generated text data. Unlike humans, who possess innate language understanding capabilities, computers process text primarily as sequences of characters or tokens. This leads to the loss of semantic and contextual awareness required to grasp human language's intricate nuances and meaning. Open information extraction (OIE) [1, 2] plays a vital role in processing and understanding large volumes of textual data. It involves identifying and extracting facts from unstructured or semi-structured sources, transforming them into structured representations without relying on predefined templates or domain-specific ontologies. Consider the example "U. S. Treasury Secretary Janet Yellen visits Beijing and plans to tell Chinese officials that Washington wants healthy economic competition." an OIE system can extract facts expressed as triples in the form $< arg1, rel, arg2 >$ (as shown in Fig. 1). In these triples, the arguments $arg1$ and $arg2$ represent entities what revolved in the facts, and the $rel$ means the relationship between the arguments. Triples are not the only way to express facts; some OIE systems extract N-ary facts from text [3]. OIE systems have been widely utilized by downstream applications, such as knowledge graph population [4, 5, 6], question answering [7], search engines [8], et al.

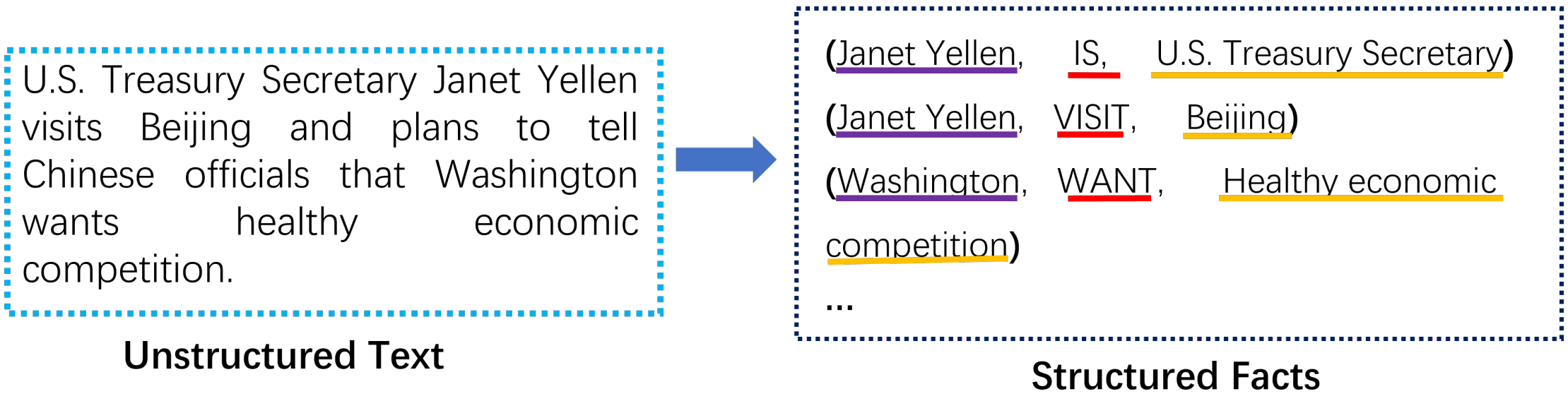

**Fig. 1** OIE systems transform unstructured text into structured expressions of facts.

There are two kinds of models to conduct OIE: linguistic grammar-oriented models and learning method-oriented models. Linguistic grammar-oriented models involve the development of handcrafted patterns or rules to identify and extract facts from text. These models rely on linguistic patterns, syntactic structures, and domain-specific heuristics to extract structured tuples representing relationships between entities and their attributes. Learning method-oriented models in OIE utilize machine learning techniques, mainly supervised approaches, to build OIE models. These models aim to capture the underlying patterns and structures in the data to generalize information extraction beyond predefined rules. The utilization of deep learning techniques in OIE has significantly advanced the field. Both kinds of OIE models have strengths and weaknesses. It seems that continued research and innovations in deep learning methods hold the potential to enhance the performance and applicability of OIE systems.

However, they often require large amounts of annotated data for training. The availability of the annotated data, particularly in non-English languages, is a significant challenge. Moreover, despite significant advancements in OIE models in recent years, the current state-of-the-art remains relatively low in overall performance and capabilities [9, 10, 12, 12]. Extracting information from Chinese text presents unique challenges due to the inherent characteristics of the Chinese language, including its rich linguistic structures, ambiguous syntax, and complex semantic nuances [13]. Consequently, developing effective OIE systems for Chinese presents a considerably more intricate and challenging task.

This paper presents a novel model, APRCOIE (Auto Patterns Recognition for Chinese Open Information Extraction), which establishes a new state-of-the-art in the field. The design of APRCOIE is based on the observations: (1) The patterns that combine lexical and syntactic features play a pivotal role in governing Chinese text; (2) These patterns cannot be effectively obtained through traditional hand-crafted approaches. In other words, while patterns exist and can be influential, they cannot be reliably captured through manual rule engineering. Based on these two observations, we designed the APRCOIE system, which comprises three components: the representation, automatic pattern acquisition, and extraction components. The representation component serves as the foundation, integrating all information involved in Chinese grammar and representing it. Its primary elements encompass part-of-speech, dependency structure, and lexical knowledge. Based on a few annotated data, the automatic pattern acquisition component generates many meaningful triple representation patterns in an automated manner. It organizes these patterns to construct a pattern library. The Extraction Component, relying on the extensive pattern repository, facilitates efficient triple extraction through a two-step process. To expedite the extraction process, we adopt a tensor computing approach, which allows for efficient computation and accelerated processing of the extraction task. We compare the performance of APRCOIE against other approaches. The experimental results demonstrate that APRCOIE achieves a new state-of-the-art in Chinese OIE.

The main contributions of this paper are twofold. (1) We propose a method of automated patterns extraction for Chinese OIE; (2) We design a schema based on tensor computing to conduct extraction with a large number of patterns efficiently; (3) We propose the APRCOIE model, which pushes the boundaries of performance of Chinese OIE systems, the model is a low resource demanding system and then can be easily deployed in real applications. (4) we manually annotate a dataset for the Chinese OIE task. The dataset contains about 8000 samples containing 12000 triples. Moreover, we annotate a dataset specific to the Chinese nominal attributes Extraction task. The code of APRCOIE and the annotated Chinese OIE dataset can be freely accessed on GitHub (https://github.com/jialin666/APRCOIE_v1).

The remainder of this paper is organized as follows: Section 2 reviews related work in the field of OIE. Section 3 presents the methodology and design principles of the APRCOIE model. In Section 4, we describe the experimental setup and evaluate the performance of APRCOIE using various metrics. Finally, in Section 5, we conclude and outline potential future directions for research in Chinese OIE.

## 2. Related work

We categorize OIE models into two main types: linguistic grammar-oriented and learning method-oriented. We proceed with reviewing these two types of models in the following.

### 2.1 Linguistic grammar-oriented models

Linguistic grammar-based models emphasize the pivotal role of linguistic grammar in model design. During the early stages of OIE modeling, the limited availability of annotated data and robust machine-learning techniques led to the emergence of linguistic grammar-oriented approaches as a prevalent option. There are various external forms of this model, including rule-based models, which directly design extraction rules depending on the linguistic grammar, and clause-based models, which usually reshape the input sentence structure according to the linguistic grammar and make it easier to conduct the extraction procedures.

Rule-based models operate based on predefined extraction rules. These rules primarily rely on manual design and require substantial linguistic knowledge. Several kinds of linguistic information can be used to construct extraction rules. Initial models utilized more basic syntactic cues like POS tags or noun phrase fragments [14, 15, 16]. Subsequent models employed deeper syntactic information, such as constituency information [17] and dependency structures [18, 19, 20, 21]. In addition to syntactic information, some models also leverage semantic information to define extraction rules [22, 23, 24]. To improve the performance, clause-based models [25, 26] incorporate a sentence transforming processing into the whole modes. The additional processing changes the complex input sentences into simple ones, thereby improving the accuracy of the models.

The linguistic grammar-oriented models are easy to implement and have low resource demands. Usually, they are characterized by their straightforwardness in understanding what these models can extract, what they cannot, and why erroneous extractions might occur. Nevertheless, there are some drawbacks to this kind of model. They are very time-consuming, which makes the models unable to be fully designed to deal with the vast array of linguistic nuances and variations. Consequently, the performance of the models might decline when dealing with unconventional language patterns or evolving language use. Additionally, an inherent challenge lies in the propagation of errors from natural language processing systems, such as POS taggers and dependency parsing tools, which can negatively impact the accuracy of the models.

## 2.2 Learning method-oriented models

Learning method-oriented models [27, 28] refer to the models that leverage machine learning methods to conduct extraction directly or indirectly. The first model, TEXTRUNNER [1], uses a self-supervised learning approach; WOE [29] and OLLIE [30] use the idea of bootstrap learning to generate training data. ReNoun [8] learns to extract noun-mediate relations with the help of distant supervision. Presently, thanks to the release of large-scale OIE datasets such as OIE2016 [9], CaRB [10], DuIE[31]et al, neural network models [32, 33] predominate the design of OIE models [34, 35, 36, 37]. When designing specific models, some treat the OIE problem as a sequence tagging task [38, 39, 40], where some tokens are tagged to signify their role in triples. In contrast, others perceive it as a generation task [41] involving the generation of entire triples. Some models incorporate rule-based methods into neural networks to enhance model performance [42, 43].

Both sequence tagging and generation approaches have their merits and demerits. Adopting neural networks diminishes the need for laborious feature engineering and minimizes the requirement for in-depth linguistic knowledge. This trend marks a central focus of current OIE research. Nonetheless, the most significant hurdle of this approach lies in its reliance on extensive annotated data. This concern is especially pronounced in the context of non-English OIE systems. Moreover, the opaque nature of neural networks can complicate model improvement efforts. Notably, recent benchmark assessments [12] reveal that, compared to rule-based methods, neural network-based methodologies have not distinctly showcased superior performance.

### 2.3 summary

Compared to the flourishing development of OIE systems for English, such as OLLIE [30], ClausIE [25], EXEMPLAR [44], Stanford OIE [26], MinIE [23], Graphene [24], et.al, OIE systems targeting other languages are notably scarce. In contrast to the diverse landscape of OIE systems available for English, there are few readily accessible OIE systems tailored for Chinese text data.

## 3. The proposed algorithm

### 3.1 Representation of pattern

In the initial part of the model, we focus on defining what constitutes a pattern for Chinese OIE. We construct a tree-like structure closely resembling the dependency tree. This pattern design comprises three essential pieces of information: dependencies, part of speech (POS) tags, and a few words that are not necessary for every pattern. In Fig. 2, two examples of patterns without words and the mapping way of forming triples are included. In the patterns, dependencies are responsible for capturing the overall syntactic structure of the text, and POS tags are used to determine what kinds of words can serve as valid tokens within the patterns. To ensure a comprehensive representation of linguistic patterns in Chinese text, we utilize 14 different dependency types and 29 distinct POS tags. This extensive set of dependencies and POS tags provides diverse forms and variations for constructing patterns. As a result, the richness in pattern forms allows our algorithm to extract structured information accurately and effectively from various sentence structures and language usages. Every pattern is accompanied by a form of triple (as shown in Fig. 2), which decides how to extract words to fill the triple slots.

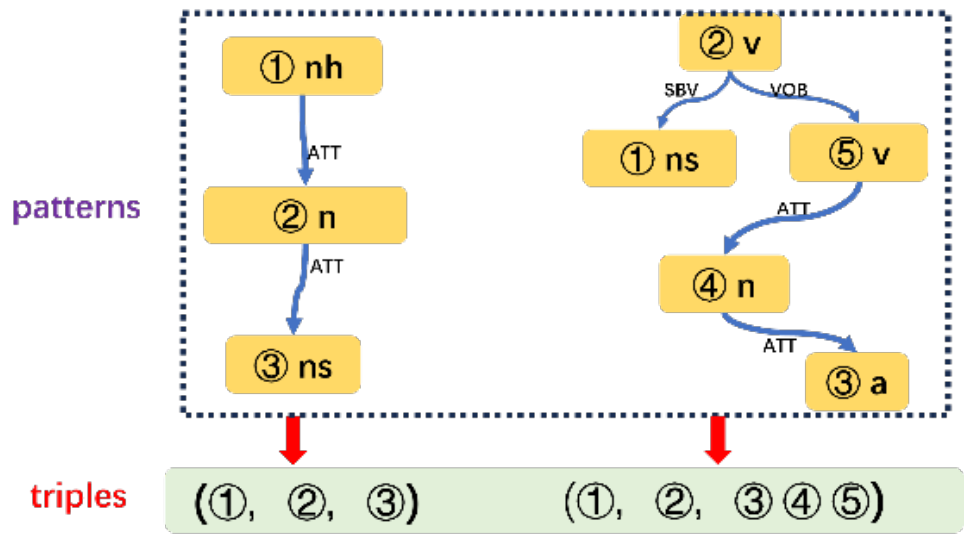

**Fig.2** The definition of patterns and the corresponding forms of triples

While dependencies and POS tags provide significant expressive power for capturing linguistic patterns, they may not always be sufficient to distinguish certain cases. We present two examples in Figure 3, where two sentences yield identical results regarding dependency parsing and POS tagging. However, despite the similarities, these sentences exhibit different forms of triples. We augment our patterns with lexical information to address the insufficiency of relying solely on dependency parsing and POS tagging. Incorporating lexical information enhances the patterns' ability to capture fine-grained nuances and disambiguate cases where dependencies and POS tags alone may not be discriminative enough. Specifically, we observed the need for additional lexical information when the POB (Preposition Object) dependency appears in the sentences. In such cases, the presence of prepositions and their associated objects can significantly impact the relationships between entities and relation phrases, leading to different triple forms. By including lexical information in the patterns, we ensure a more comprehensive and accurate representation of the linguistic patterns present in the text. The combination of dependencies, POS tags, and lexical information empowers our algorithm to effectively distinguish between subtle linguistic variations, even in cases where dependency parsing and POS tagging may yield identical results.

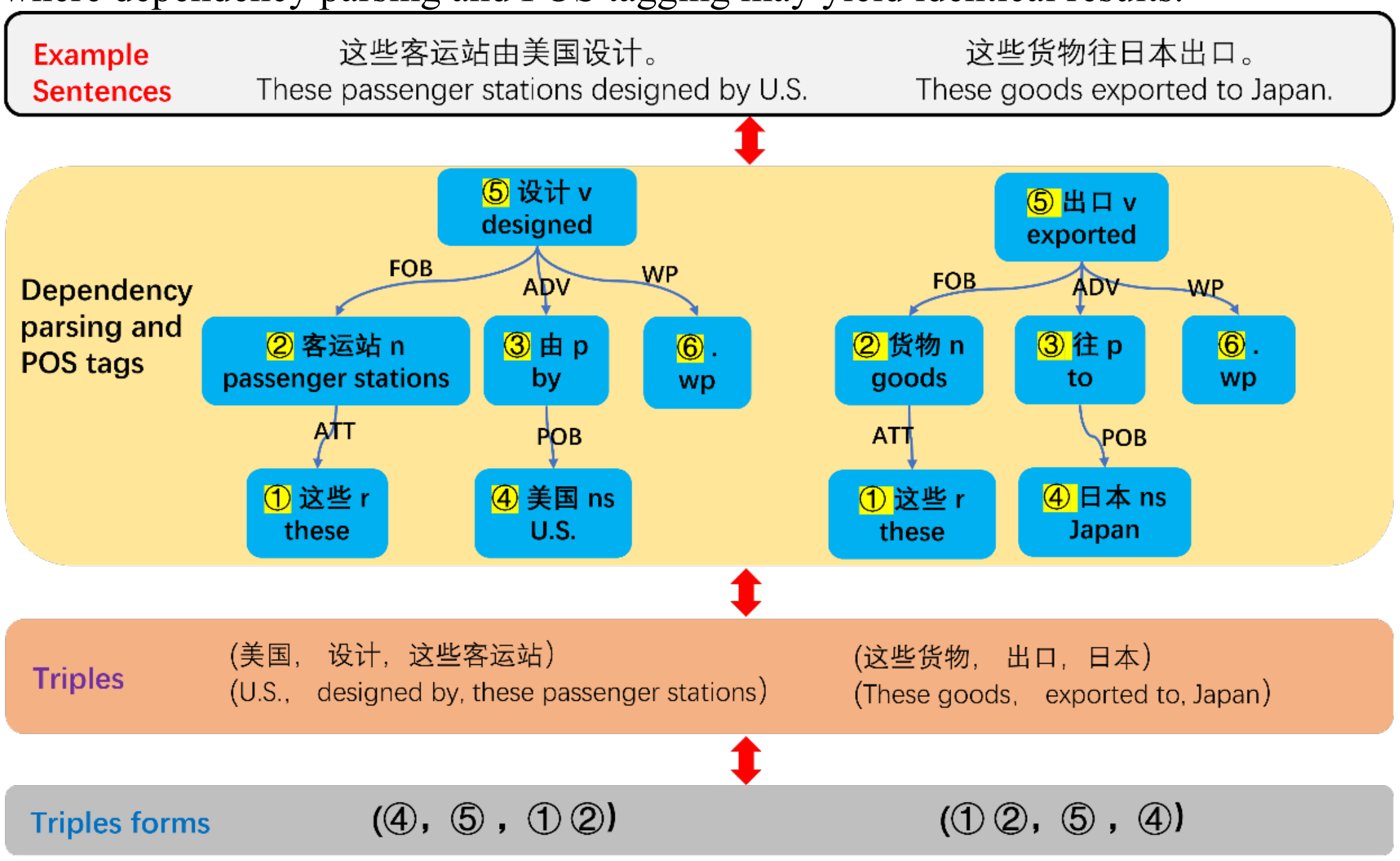

**Fig. 3** Adding lexical information to patterns is necessary in some cases.

Including lexical information in the patterns allows our algorithm to handle a broader range of linguistic phenomena and provides more reliable and robust information extraction results. However, we take a cautious approach and aim to minimize the use of lexical information in the pattern design. The reason for this stems from the vast number of words in Chinese, which can lead to excessive complexity in patterns if too many words are involved. We use lexical information only in cases with identical shallower and deeper syntactic information but different filling ways of triple's slots.

### 3.2 Automatic acquiring of patterns

In the second step of our model, we focus on generating patterns set from annotated data. To ensure the quality and reliability of the pattern set, it is crucial to have high-quality annotated data. We manually annotate a Chinese OIE dataset where most samples are chosen from the SAOKE dataset [36]. We meticulously reviewed the SAOKE dataset, identified and eliminated errors, and removed classical Chinese data that diverges from the grammar of modern Chinese. Through these efforts, we obtained a dataset comprising about 7000 instances of meticulously annotated Chinese OIE data. The details of the dataset are shown in Table 1, #A means the number of A, SVO is the structure of Subject+Verb+Object, SVOCOO means the coordinating conjunctions of SVO, Nominal represents the nominal attributes, POB is the prepositional phrase.

**Table 1.** The statistics of the annotated dataset for Chinese OIE

| #sentences | #triples | #SVO | #SVOCOO | #Nominal | #POB |
|---|---|---|---|---|---|
| 7878 | 14084 | 7511 | 3205 | 205 | 3163 |

The generation of patterns involves a mapping procedure that considers dependencies, POS tags, and lexical information. The mapping procedure is depicted in Fig. 4. It starts with inputting dependencies, POS tags, and lexical information. Then, it identifies the first common ancestor of all tokens in the annotated triple from the dependency tree (For the given annotated triple (华盛顿 (Washington), 希望(wants), 健康经济竞争(healthy economic competition)) as shown in Fig. 4, the common ancestor is 希望(wants)). Any words that are neither part of the annotated triple nor in the path from nodes in the triple to the common ancestor are eliminated from the dependency tree. The words in the rest of the dependency tree are replaced with their respective POS tags, and we will obtain the tree structures. Then, we will compare each pair of the tree structure. Suppose there are identical tree structures with different ways of arranging the tokens in triples, and prepositions appear in these tree structures. The tree structures with the preposition will be the final pattern in that case. Otherwise, the tree structures without any lexical information are the final patterns. Finally, the procedure generates a triple form that provides instructions on how to fill the triple slots.

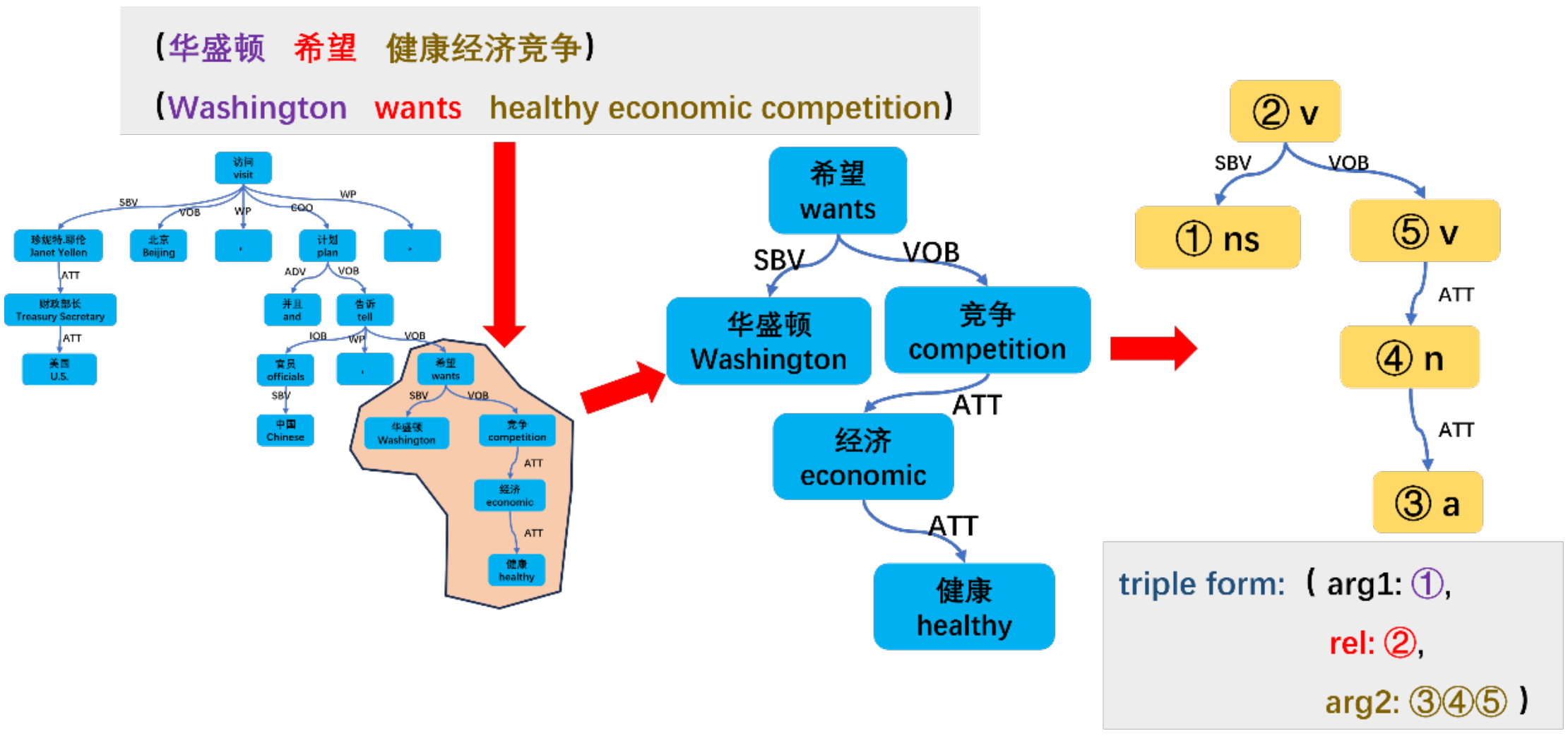

**Fig. 4** The procedure for generating patterns

We show more examples of generating patterns in Fig. 5. It notes that in the last two examples, the words 由(by) and 往(to) are involved in the pattern. These two words are not part of the triples but part of the patterns.

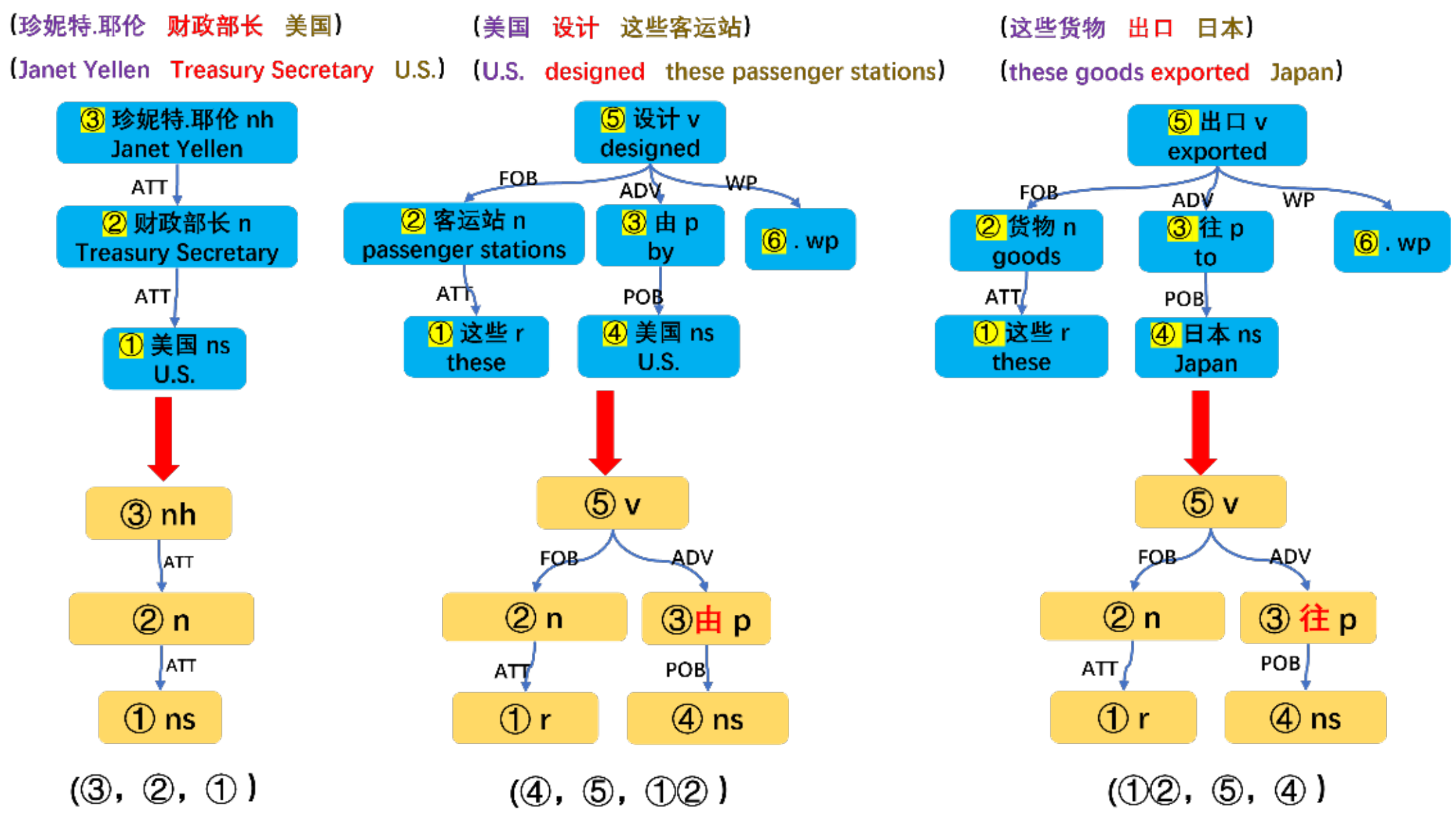

**Fig. 5** More examples of generating patterns

### 3.3 Extracting of triples

The previous phase has resulted in many patterns that must be utilized in the extracting step. Given the massive scale of text data typically involved in OIE tasks, the efficiency of the triple extraction process becomes a critical consideration. In this part, we present a highly efficient algorithm specifically designed for extracting triples based on a large scale of patterns. The two-step algorithm efficiently identifies and extracts relevant triples based on a large number of generated patterns. The first step serves as a preliminary filter, while the second step carries out the actual triple extraction using the filtered patterns.

To streamline the triple extraction process and handle the large pattern volume, we leverage matrix computing in the preliminary filter of patterns. It first transforms every pattern into a matrix form called a pattern matrix. Every sample in the annotated data generates a pattern; then, the pattern will be represented by the pattern matrix. The construction of a pattern matrix is based on the linguistic elements present in the pattern, where each row or column corresponds to a specific POS tag, representing a word in the pattern. Each kind of dependency has a unique positive value; 0 means no dependency value. We utilize the 14 different POS tags and 29 types of dependencies []. Then, every pattern is represented by a 14x14 pattern matrix with 30 kinds of values; the whole procedure is shown in Fig. 6. The target sentence is also represented by a matrix which is generated in the same way and called sentence matrix (as shown in Fig. 7). The preliminary filtering is conducted by the subtraction between these two kinds of matrices: First all pattern matrices are combined to form a 3-D tensor which called pattern tensor. Then, the target sentence matrix is broadcasted to the size of the pattern tensor to create the sentence tensor. Then, the sentence tensor subtracts the pattern tensor. The result tensor filters the patterns whose matrix contains negative numbers.

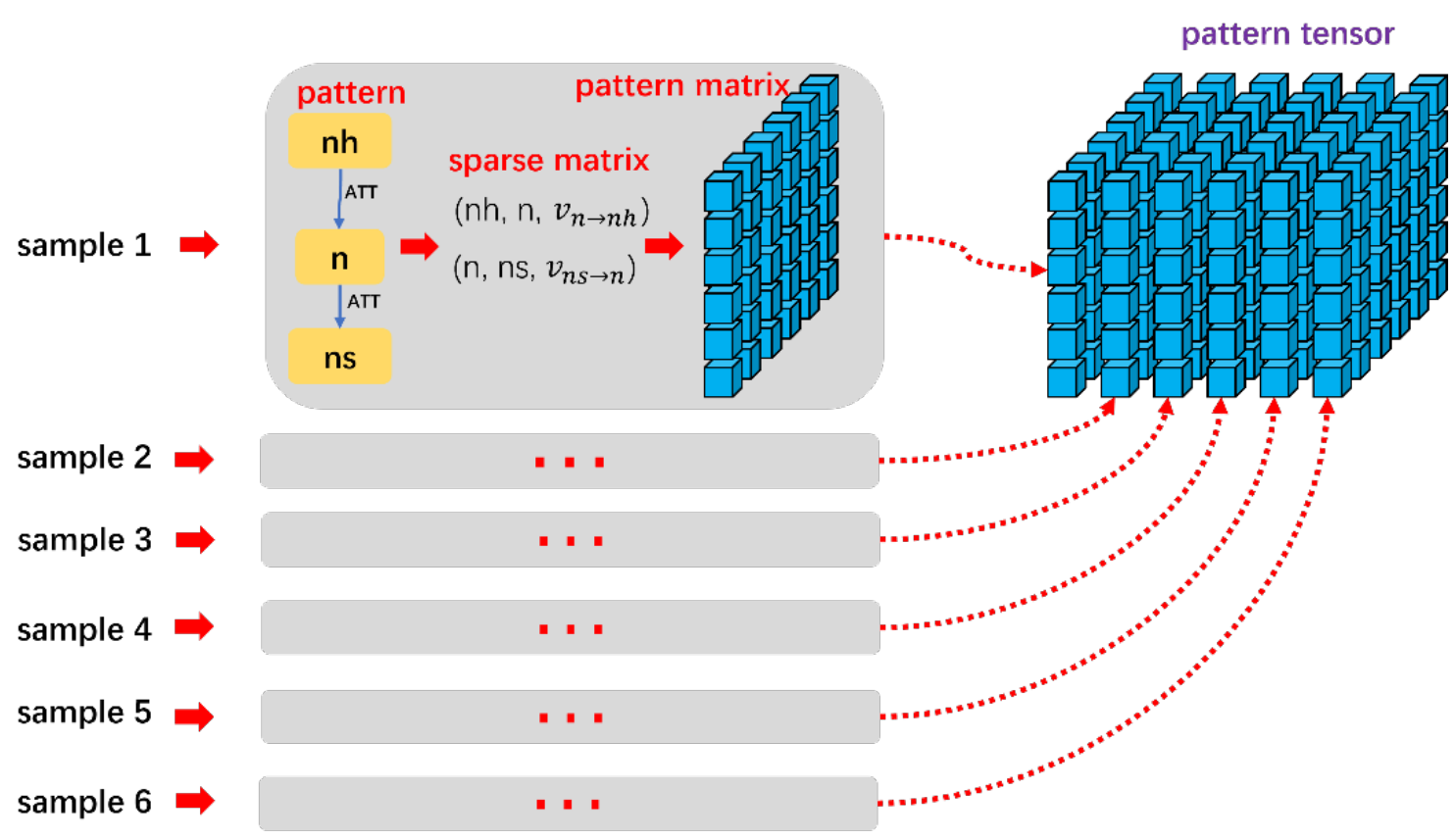

**Fig. 6** The generation of pattern tensor

In the second step, we perform the actual extraction of triples using the patterns that survived the preliminary filter. Each retained pattern is processed individually, and the algorithm extracts triples from the corresponding segments of text data. Focusing only on the relevant patterns identified in the first step ensures that the triple extraction procedure remains targeted and precise. This step leverages the patterns' structure and the matching process to capture and extract structured information from the text accurately.

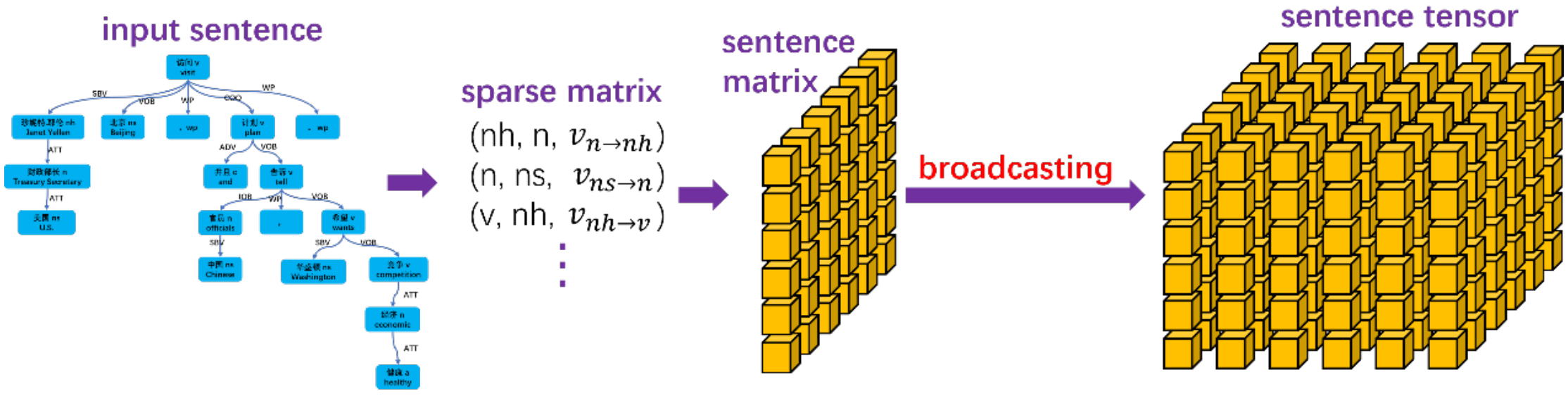

**Fig.** 7 The generation of target sentence tensor

### 3.4 Post-processing for triples

In the post-processing step of our algorithm, we address the issue of incomplete triples that may arise during the argument and relation extraction stages. These incompletenesses are primarily observed in two cases: 1) when existing patterns do not cover specific noun modifiers or quantifiers during argument extraction and 2) when relation extraction requires the inclusion of modifiers with negative meanings as part of relation phrases. To mitigate these issues and enhance the completeness of the extracted triples, we introduce a post-processing step that focuses on the expansion of triples. This step involves refining the initial triples by incorporating additional relevant information from the surrounding context of the sentence.

For argument extraction, when the existing patterns miss specific noun modifiers or quantifiers, the post-processing step analyzes the sentence context to identify potential missing elements that should be part of the argument. By leveraging contextual cues, such as nearby words and syntactic relationships, we expand the extracted arguments to ensure a more comprehensive representation of the entities involved in the triple. Similarly, during relation extraction, if specific modifiers with negative meanings are crucial to forming accurate relation phrases, the post-processing step incorporates these modifiers into the relation to capture the intended meaning more precisely.

The post-processing step is carefully designed to strike a balance between expanding the extracted triples for completeness while avoiding overfitting or introducing irrelevant information. By leveraging contextual information and linguistic cues, we ensure the expanded triples remain coherent and contextually relevant.

## 4. Experiments

In this part, we present a comprehensive evaluation of APRCOIE. We compare several algorithms in a fact-based benchmark dataset [12] and a nominal attributes dataset.

### 4.1 Training data

We acknowledge the crucial role of high-quality annotated data in developing the OIE field. However, obtaining such data, especially for Chinese OIE in the (subject, predicate, object) triples form, is a significant challenge due to its scarcity. While there are a few datasets for Chinese OIE, many focus on n-ary extracting tasks or unique for span models or other cases, lacking the specific (subject, predicate, object) triples form that is the most common case in OIE.

To address this limitation, we manually collected sentences from various sources, including the SAOKE dataset [36] and other relevant materials. Our efforts created a Chinese OIE dataset specifically tailored for triple-form extracting. The dataset comprises 8000 sentences and 14000 facts in the (subject, predicate, object) triples form. To ensure the data's reliability and accuracy, all samples in the dataset have undergone meticulous manual annotation by two annotators to enhance the quality of the annotations. We show the details in Table 1。

By releasing this new Chinese OIE dataset, we aim to fill the gap in the availability of high-quality data for (subject, predicate, object) triples form extraction, enabling researchers to evaluate and advance their algorithms in this critical area.

**4.2 Evaluation of Chinese open IE systems**

Evaluating the performance of various Chinese OIE models is a challenging endeavor due to the inherent underspecification of the task and the lack of a definitive gold standard. The subjective nature of determining whether extracted triples are acceptable or not introduces inconsistency in the evaluation process. Recognizing these challenges, we adopt a recently published benchmark BenchIE [12] to evaluate various models. The new benchmark BenchIE exhausts all explicit triples and clustered triples with the same fact into a fact synset. Unlike previous benchmarks, BenchIE evaluates model performance based on the alignment of extracted triples with fact synsets rather than the gold triples. This inherent nature renders it the most objective benchmark for OIE tasks.

**Table 2**. Performance of APRCOIE and baseline models on the BenchIE dataset

| | BenchIE dataset | | |
|---|---|---|---|
| **Model** | **Precision** | **Recall** | **F1** |
| DSNF | 0.06 | 0.069 | 0.065 |
| ZORE | 0.097 | 0.064 | 0.077 |
| M2OIE | 0.176 | 0.103 | 0.130 |
| APRCOIE | **0.45** | **0.154** | **0.226** |

We assessed the performance of our proposed model in comparison with three other prominent models: the DSNF model [20], the ZORE model [19], and the M2OIE [39] model. The LTP system has been used as the natural language tool for POS tagging and dependency parsing [45]. The evaluation used standard metrics, including precision, recall, and the F1 score. The performance results for each model are presented in Table 2, clearly demonstrating our model's superior performance across all metrics. Specifically, there was an improvement of approximately 10% in the F1 score, 5% in the recall score, and a remarkable 27.4% boost in the precision score.

Recognizing that the benchmark BenchIE lacked the inclusion of nominal attribute extraction, which is especially useful in searching engines and question answers [2014-EMNLP-1], we conducted experiments dedicated to extracting nominal attributes. We assembled a specialized dataset specifically curated for nominal attribute extraction by selecting all relevant samples from our annotated dataset. In total, the dataset comprises approximately 200 samples exemplifying nominal attributes. We take 50 samples as test data and the rest as training data.

We compare our model with the model of DSNF afourbilities to accurately extract nominal attributes. The specific definitions of these metrics are as follows. We use $G = \{G_1, G_2, \cdots, G_n\}$ to denote the gold triples, $G_i = \{g_{i1}, \cdots, g_{ih}\}$ is the gold triples for the ith sample, $g_{ij} = [g_{ij}^{arg1},\ g_{ij}^{rel},\ g_{ij}^{arg2}]$ is the jth gold triple of the ith sample. Let $E = \{E_1,\ E_2,\ \cdots,\ E_n\}$ is the model's extractions, $E_i = \{e_{i1}, \cdots, e_{ik}\}$ is the extraction result of the ith sample, $e_{ic} = [e_{ic}^{arg1},\ e_{ic}^{rel},\ e_{ic}^{arg2}]$ is the cth extracted triple of the ith sample, where $e_{ic}^{arg1}$ is the first argument, $e_{ic}^{rel}$ is the relation, $e_{ic}^{arg2}$ is the second argument. Let $|e_{ic}^{arg1} \cap g_{ij}^{arg1}|$ be the number of common characters between $e_{ic}^{arg1}$ and $g_{ij}^{arg1}$, $|e_{ic}|$ be the number of characters in triple $e_{ic}$. Then for a triple pair to a sample, we have the scores:

$$precision(e_{ic}, g_{ij}) = \frac{|e_{ic}^{arg1} \cap g_{ij}^{arg1}| + |e_{ic}^{rel} \cap g_{ij}^{rel}| + |e_{ic}^{arg2} \cap g_{ij}^{arg2}|}{|e_{ic}|}$$

$$recall(e_{ic}, g_{ij}) = \frac{|e_{ic}^{arg1} \cap g_{ij}^{arg1}| + |e_{ic}^{rel} \cap g_{ij}^{rel}| + |e_{ic}^{arg2} \cap g_{ij}^{arg2}|}{|g_{ij}|}$$

$$F_1(e_{ic}, g_{ij}) = \frac{2\ \cdot precision(e_{ic}, g_{ij})\ \cdot recall(e_{ic}, g_{ij})}{precision(e_{ic}, g_{ij}) + recall(e_{ic}, g_{ij})}$$

To compute the overall performance metrics, we greedily match extracted triples in $E_i$ with reference ones in $G_i$ at each time. For triple $e_{ic}$ in $E_i$, we denote the matching gold in $G_i$ as $g_{ic^*}$, and conversely $e_{ij^*}$ matches $g_{ij}$. Hence the overall performance metrics defined as follow:

$$cw_{ic} = |e_{ic}^{arg1} \cap g_{ic^*}^{arg1}| + |e_{ic}^{rel} \cap g_{ic^*}^{rel}| + |e_{ic}^{arg2} \cap g_{ic^*}^{arg2}|$$

$$cw_{ij} = \left|e_{ij^*}^{arg1} \cap g_{ij}^{arg1}\right| + |e_{ij^*}^{rel} \cap g_{ij}^{rel}| + |e_{ij^*}^{arg2} \cap g_{ij}^{arg2}|$$

$$precision = \frac{\sum_i^n \sum_c^k cw_{ic}}{\sum_i^n \sum_c^k |e_{ic}|}$$

$$recall = \frac{\sum_i^n \sum_c^h cw_{ij}}{\sum_i^n \sum_c^h |g_{ij}|}$$

$$F_1 = \frac{2 \cdot precision \cdot recall}{precision + recall}$$

The results of the experiment on nominal attribute extraction are presented in Table 3. Notably, in comparison to the previous experiments conducted on the BenchIE dataset, the performance outcomes of the compared models exhibit marked improvement. However, our model consistently outperforms the competing approaches across all metrics. The metrics' values achieved by our model are demonstrably higher, signifying a substantial performance advantage over the other models.

**Table 3**. Performance of MY and baseline models on the nominal attributes dataset

| | **Nominal attributes dataset** | | |
|---|---|---|---|
| **Model** | **Precision** | **Recall** | **F1** |
| DSNF | 0.302 | 0.284 | 0.291 |
| ZORE | 0.401 | 0.307 | 0.348 |
| APRCOIE | **0.921** | **0.846** | **0.882** |

**4.3 Analysis**

The above experiments show that our algorithm has achieved significant improvements across all performance metrics. The reason lies in the innovation we've applied from both effectiveness and efficiency perspectives.

Firstly, a distinctive element that sets our algorithm apart lies in its strategic departure from conventional rule-based models. Previous research endeavors often aimed to distill a handful of general rules for extraction, designing algorithms based on these rules. However, the dynamic nature of Chinese linguistic patterns often rendered these general rules ineffective, leading to incoherent and uninformative extractions. In contrast, our model operates with a different philosophy, opting for an expansive approach. It generates many extraction patterns, intentionally sidestepping the expectation that each rule possesses inferential capability. This strategy embraces the diversity of linguistic patterns present in Chinese text. During the extraction process, only the patterns that exhibit precise matches with the target are invoked, ensuring that coherent and relevant extractions are produced. This innovative strategy accounts for our algorithm's capability to achieve the state of the art.

Secondly, as to why the exact matching strategy can function efficiently, credit can be attributed to the preliminary filter procedure. Conducting a precise matching of a vast array of patterns with the target sentence can be time-consuming. However, the preliminary filter process effectively addresses this challenge by significantly reducing the number of patterns engaged during extraction. By filtering out the majority of patterns, the preliminary filter optimizes the extraction process. Consequently, the final number of rules that are actually involved in the extraction remains minimal, thereby streamlining the computational demands and enhancing the efficiency of our algorithm.

## 5. Conclusion and Future Work

In this paper, we introduce a novel Chinese OIE model APRCOIE that not only addresses the inherent complexities of Chinese linguistic patterns but also pushes the boundaries of performance of Chinese OIE. The APRCOIE is a low-resource demanding model and can be easily deployed in real applications. We make the APRCOIE system, together with a re-annotated dataset freely available at GitHub.

The experimental results provide a revealing insight into the dynamics of model performance. Interestingly, prior to the introduction of our model, it is evident that neural network-based OIE models were consistently outperforming rule-based counterparts whatever in English or other languages. However, the introduction of APRCOIE marks a notable turning point in this trajectory. Through our model, we demonstrate that rule-based models retain the potential to achieve state of the art.

A critical insight from our error analysis points towards the potential avenues for refining our model. The avenue of semantic understanding emerges as a pivotal direction for future enhancements. While our rule-based approach has demonstrated remarkable prowess, the incorporation of semantic information could potentially propel our model to even greater heights of accuracy and coherence.

**Declaration of Competing Interest**

The authors declare there is no conflict interest.

**Acknowledgements**

The authors would like to thank the anonymous referees for their helpful comments in improving the presentation of the paper. This work is supported by the Educational Commission of Jiangxi Province (Project No. GJJ2200502). This work is supported in part by the National Social Science Fund, China, under Weiying Ping (Project No.

20&ZD131), under Chunhai Tao (Project No. 21&ZD150). Lu's research is partially supported by a Discovery Grant (RG/PIN06466-2018) from Natural Sciences and Engineering Research Council (NSERC) of Canada.